\documentclass{article}
\usepackage{spconf,amsmath,graphicx}
\usepackage{url}            
\usepackage{booktabs}       
\usepackage{amsfonts}       
\usepackage{nicefrac}       
\usepackage{microtype}      
\usepackage{xcolor}         
\usepackage{graphicx}
\usepackage{multirow}
\usepackage{bbding}
\usepackage{enumitem}
\usepackage[colorlinks,linkcolor=black,anchorcolor=black,citecolor=black]{hyperref}

\title{LEFormer: A Hybrid CNN-Transformer Architecture for Accurate Lake Extraction from Remote Sensing Imagery}
\name{Ben Chen$^{1, }$$^*$, Xuechao Zou$^{1, }$$^*$, Yu Zhang$^{1}$, Jiayu Li$^{1}$, Kai Li$^{2}$, Junliang Xing$^{2}$, Pin Tao$^{1,2}$$^\dagger$\thanks{$^*$These authors contributed equally.}\thanks{$^\dagger$Corresponding author.}}
\address{
$^1$Qinghai University, Department of Computer Technology and Applications, Xining, China\\
$^2$Tsinghua University, Department of Computer Science  and Technology, Beijing, China
}

\begin{document}
%
\maketitle
\begin{abstract}
  Lake extraction from remote sensing images is challenging due to the complex lake shapes and inherent data noises. Existing methods suffer from blurred segmentation boundaries and poor foreground modeling. This paper proposes a hybrid CNN-Transformer architecture, called LEFormer, for accurate lake extraction. LEFormer contains three main modules: CNN encoder, Transformer encoder, and cross-encoder fusion. The CNN encoder effectively recovers local spatial information and improves fine-scale details. Simultaneously, the Transformer encoder captures long-range dependencies between sequences of any length, allowing them to obtain global features and context information. The cross-encoder fusion module integrates the local and global features to improve mask prediction. Experimental results show that LEFormer consistently achieves state-of-the-art performance and efficiency on the Surface Water and the Qinghai-Tibet Plateau Lake datasets. Specifically, LEFormer achieves 90.86\% and 97.42\% mIoU on two datasets with a parameter count of 3.61M, respectively, while being 20$\times$ minor than the previous best lake extraction method. The source code is available at \url{https://github.com/BastianChen/LEFormer}.
\end{abstract}
\section{Introduction}
\label{sec:intro}

\begin{figure}
  \centering
  \noindent\includegraphics[width=0.8\linewidth]{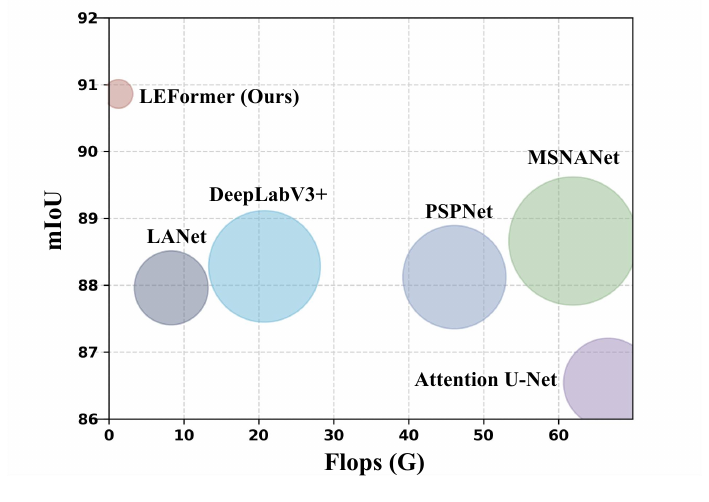}
  \caption{Performance and efficiency comparison on Surface Water dataset. The circle area indicates the number of parameters.
  }
  \label{fig: bubble}
\end{figure}

\begin{figure*}
  \centering
  \noindent\includegraphics[width=0.76\linewidth]{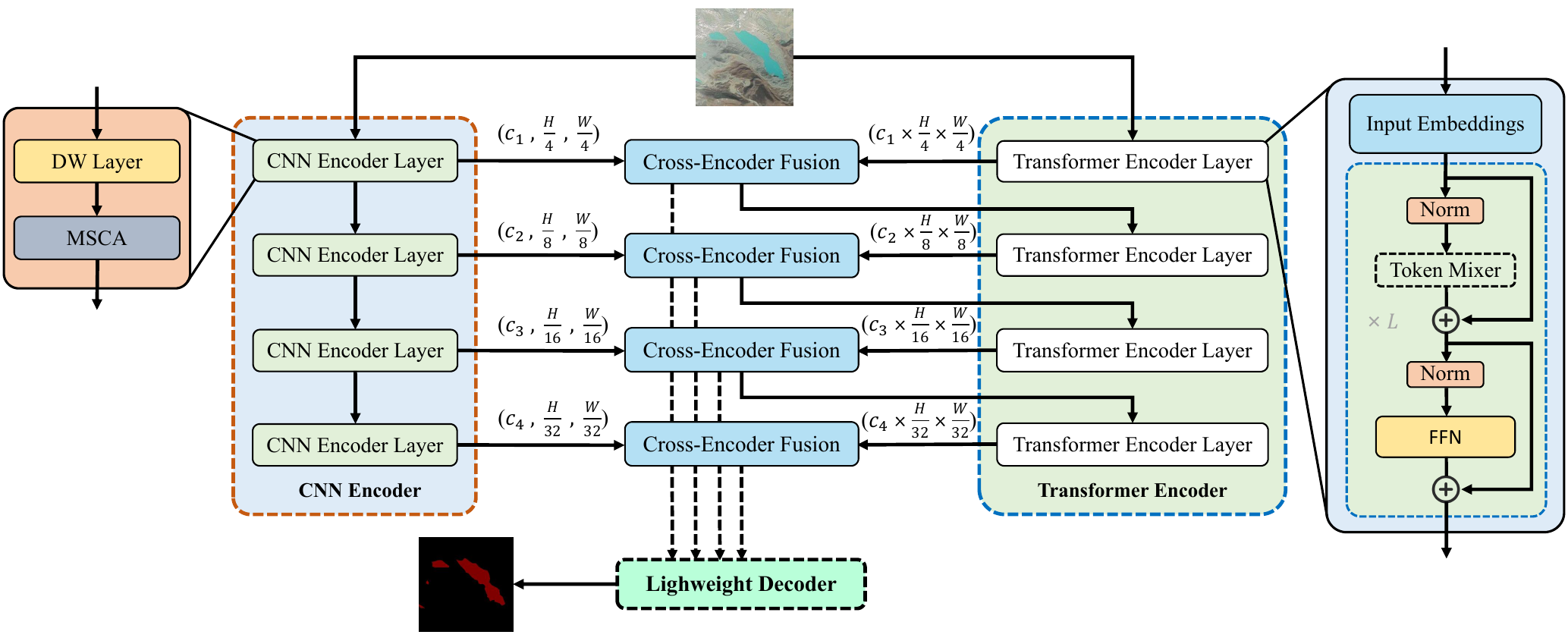}
  \caption{Overview architecture of our proposed LEFormer.
  }
  \label{fig:overall_architecture_diagram}
\end{figure*}



Lakes are crucial environmental and climatic indices that have gained focus~\cite{ClimaticIndicator}. Advancements in worldwide observation technologies and sensing equipment have made remote sensing images a common means for extracting hydrological features~\cite{HA-Net}. Automated extraction of lakes from remote sensing data is critical to track climate alterations~\cite{MSCANet}.

Lake extraction has generally been considered a semantic segmentation task. Recently, researchers have utilized deep learning methods for lake extraction, such as UDGN~\cite{UDGN}, to boost the spatial resolution of lake zones. MSLWENet~\cite{QTPL-Dataset} proposes an end-to-end multi-scale plateau lake extraction network model based on ResNet-101~\cite{ResNet} and depth-wise separable convolution~\cite{Depth-wiseConvolutionLayer}. HA-Net~\cite{HA-Net} presents a mixed-scale attention mechanism, while MSNANet~\cite{MSNANet} employs an encoder-decoder backbone to improve feature representation and achieve superior lake extraction outcomes. Nevertheless, the extraction of lakes remains challenging due to high interclass heterogeneity and complex background information, such as snow, glaciers, and mountains, which introduce contextual ambiguity and pose additional extraction challenges.

To address the limitations of existing models, we propose an efficient architecture (LEFormer) for extracting lakes from remote sensing imagery by combining CNN and Transformer architectures. Our architecture leverages CNN to extract local features and Transformer to capture global features. The cross-encoder fusion module fuses the local and global features extracted by the CNN and Transformer into a unified feature used as input to the generated lake mask. These modules achieve high accuracy and low computational cost with a lightweight network structure, as shown in Fig. \ref{fig: bubble}. The main contributions of our architecture are summarized as follows:
\begin{itemize}
\item We introduce LEFormer, a novel architecture for lake extraction. LEFormer combines CNN and Transformer to capture local and global features and employs cross-encoder fusion modules to improve mask prediction.
\item We design a CNN encoder with multi-scale spatial-channel attention (MSCA) to extract precise and detailed local spatial information.
\item We develop a lightweight Transformer encoder,
reducing the computational and parameter demands of the model while maintaining high performance.
\end{itemize}

\section{METHODOLOGY}

In this study, we propose the LEFormer, a novel architecture for high-performance lake extraction. Illustrated in Fig. \ref{fig:overall_architecture_diagram}, LEFormer contains three main modules: (1) CNN Encoder (CE) for precise local spatial information extraction; (2) Transformer Encoder (TE) for capturing long-range dependencies and global context; (3) lightweight Cross-Encoder Fusion (CEF) module for integrating features from TE and CE.

\subsection{CNN Encoder (CE)}

\begin{figure*}[t]
  \centering
  \noindent\includegraphics[width=0.75\linewidth]{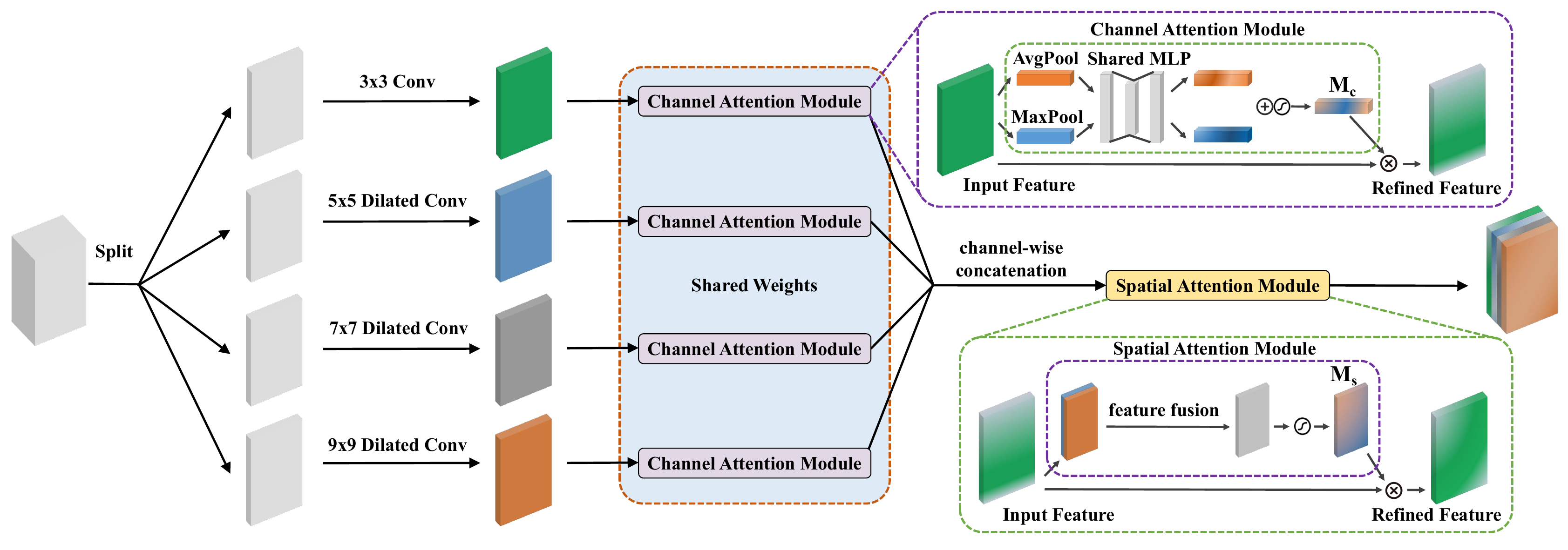}
  \caption{The diagram depicts the structural layout of the MSCA. }
  \label{fig:MSCA}
\end{figure*}

Our proposed CE is based on a hierarchical structure consisting of multiple stacked CE layers. Each CE Layer comprises a depth-wise separable (DW) layer~\cite{Depth-wiseConvolutionLayer} and an MSCA, facilitating improved multi-scale feature extraction. We use the DW Layer as down-sampling stages to down-sample the input image to sizes of $\frac{H}{4} \times \frac{W}{4}$, $\frac{H}{8} \times \frac{W}{8}$, $\frac{H}{16} \times \frac{W}{16}$, and $\frac{H}{32} \times \frac{W}{32}$, where $H$ and $W$ denote the height and width of the input image. Each down-sampling stage contains a down-sampling block with a convolutional layer (detailed settings are listed in Table \ref{table:network_settings}) and a GELU activation function~\cite{GELUS}.

In addition, we integrate an MSCA that combines the benefits of dilated convolutions~\cite{dilated_convolutions} and CBAM~\cite{CBAM} for enhanced multi-scale feature extraction. Dilated convolutions can improve segmenting objects or features of different scales by adjusting the kernel dilation rate. We first utilize dilated convolutions with 1, 2, 3, and 4 dilation rates to generate multi-scale feature maps. Next, we apply the CBAM layer to compute attention weighting for the extracted multi-scale feature maps and multiply it with the input multi-scale feature maps to obtain the final features. The CBAM layer employs channel and spatial attention mechanisms that dynamically allocate weights to the features of the captured lake mask, effectively enhancing the model's feature representation and discrimination capability. This combination of techniques enables efficient aggregation of relevant information at multiple scales while minimizing computational cost. The structure of the MSCA is illustrated in Fig. \ref{fig:MSCA}.

\begin{table}[t]
\small
\centering
\setlength{\tabcolsep}{2pt}
\caption{The detailed settings of our LEFormer, including patch size ($K_i$), stride ($S_i$), padding size ($P_i$), channel number ($C_i$), reduction ratio ($R_i$), head number ($N_i$)  and number of encoder blocks ($L_i$) in Stage i.}
\begin{tabular}{c|c|c|c}
\hline
Stage & Output Size                        & CNN Encoder                                                              & Transformer Encoder \\ \hline
1     & $\frac{H}{4} \times \frac{W}{4}$   & \begin{tabular}[c]{@{}c@{}}$K_1=7;S_1=4;$\\ $P_1=3;C_1=32$\end{tabular}  & $R_1=8;N_1=1;L_1=2$ \\ \hline
2     & $\frac{H}{8} \times \frac{W}{8}$   & \begin{tabular}[c]{@{}c@{}}$K_2=3;S_2=2;$\\ $P_2=1;C_2=64$\end{tabular}  & $R_2=4;N_2=2;L_2=2$ \\ \hline
3     & $\frac{H}{16} \times \frac{W}{16}$ & \begin{tabular}[c]{@{}c@{}}$K_3=3;S_3=2;$\\ $P_3=1;C_3=160$\end{tabular} & $R_3=2;N_3=5;L_3=2$ \\ \hline
4     & $\frac{H}{32} \times \frac{W}{32}$ & \begin{tabular}[c]{@{}c@{}}$K_4=3;S_4=2;$\\ $P_4=1;C_4=192$\end{tabular} & $R_4=1;N_4=6;L_4=3$ \\ \hline
\end{tabular}
\label{table:network_settings}
\end{table}

\subsection{Transformer Encoder (TE)}


Motivated by Transformers' success in computer vision, we propose a lightweight TE layer. ViT~\cite{ViT} lacks local continuity among patches and requires positional embedding interpolation when test resolution differs from training, reducing accuracy. To address this, we adopt Overlapped Patch Merging from SegFormer~\cite{SegFormer}, with the same configuration as CE.

FFN uses a 2D convolutional layer with a kernel size of $3\times3$ to extract positional information instead of traditional positional encoding, expressed as follows:
\begin{equation}  
\mathbf{F}_{out} = \text{MLP}(\text{GELU}(\mathrm{Conv_{3\times 3}}(\text{MLP}(\mathbf{F}_{in}))))+\mathbf{F}_{in}. \tag{1}
\label{Eq:1}
\end{equation}


\begin{figure}[t]
\centering
\begin{tabular}{cccc}
\includegraphics[width=0.21\linewidth]{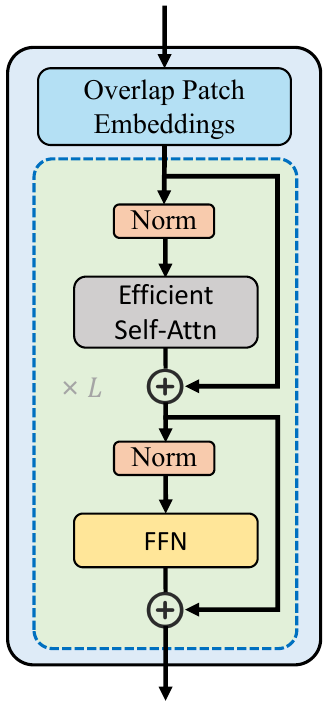} &
\includegraphics[width=0.21\linewidth]{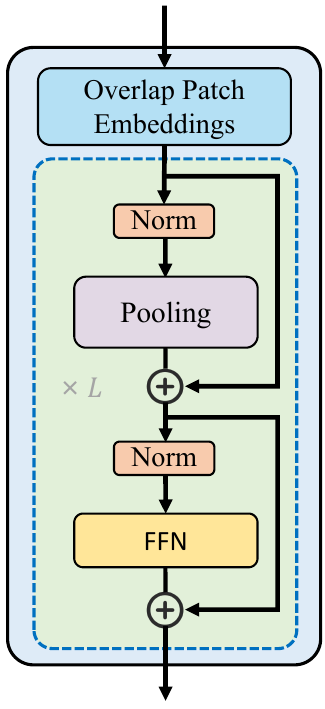} &
\includegraphics[width=0.43\linewidth]{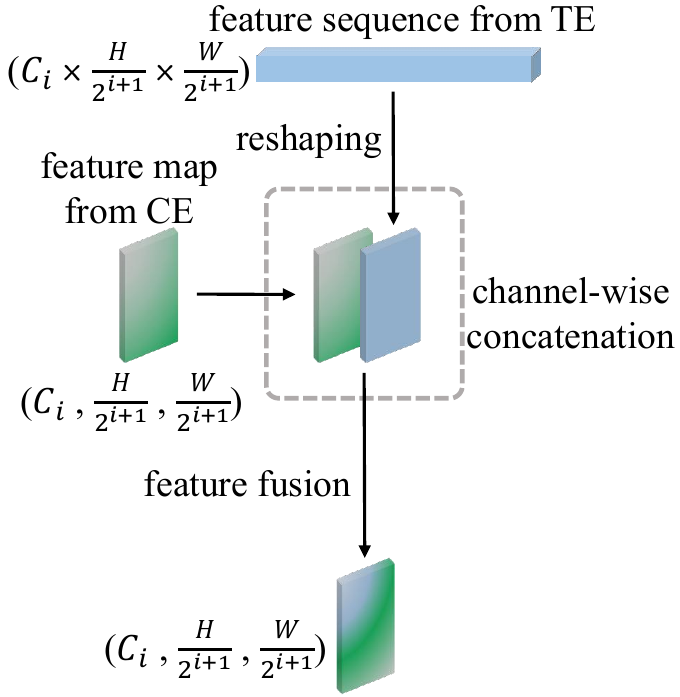} 
\\
(a) & (b) & (c) 
\end{tabular}
\caption{The diagram depicts the structural layout of three critical modules: (a) the Efficient Transformer Layer (ETL), (b) the Pooling Transformer Layer (PTL), and (c) the Cross-Encoder Fusion (CEF).}
\label{fig:transformer_encoder_layer}
\end{figure}

In addition to the above issues, the self-attention process is computationally expensive and has a complexity of $\mathcal{O}(N^2)$, making it impractical for large image resolutions. Efficient Self-Attention~\cite{SegFormer} tackles this problem by utilizing the sequence reduction process from PVT~\cite{PVT}, which reduces the sequence length using a reduction ratio $R$ as follows:

\begin{equation}\label{Eq:2}
\begin{gathered}
\mathbf{\hat{K}} = \mathrm{Conv_{R\times R}}(\text{Reshape}(C,H,W))(\mathbf{K}) \\
\mathbf{K} = \text{Reshape}\left(\frac{N}{R^2},C\right)(\mathbf{\hat{K}}), 
\end{gathered}
\tag{2}
\end{equation}
where $N$ denotes $H \times W$. Efficient Self-Attention reduces sequence $\mathbf{K}$ using a $\mathrm{Conv_{R\times R}}$ layer and $\text{Reshape}(\frac{N}{R^2}, C)(\mathbf{\hat{K}})$. It reshapes $\mathbf{K}$ into a $(C, H, W)$ tensor, applies $\mathrm{Conv_{R\times R}}$ to generate $(C, \frac{H}{R}, \frac{W}{R})$, and uses $\text{Reshape}(\frac{N}{R^2}, C)$ to reduce $\mathbf{\hat{K}}$ to $\frac{N}{R^2} \times C$. This reduces self-attention complexity from $\mathcal{O}(N^2)$ to $\mathcal{O}(\frac{N^2}{R^2})$. Overlapped Patch Merging, FFN, and Efficient Self-Attention form an ETL (Fig. \ref{fig:transformer_encoder_layer}(a)).
Inspired by PoolFormer's~\cite{PoolFormer} success in the lightweight vision transformer domain, we combine Overlapped Patch Merging, FFN, and a spatial pooling operator to form a PTL (Fig. \ref{fig:transformer_encoder_layer}(b)). This work combines ETL and PTL by designing four Encoder Layers. The first layer employs PTL, while the subsequent three layers employ ETL to extract global features while minimizing computational cost. Refer to Table \ref{table:network_settings} for more details.

\subsection{Cross-Encoder Fusion (CEF)}
We propose a lightweight layer (Fig. \ref{fig:transformer_encoder_layer}(c)) that combines local features from the CE Layer with global features from the TE layer. The CE layer's feature map has size $(C_i,\frac{H}{2^{i+1}},\frac{W}{2^{i+1}})$, while the TE layer's feature sequence has size $(C_i \times \frac{H}{2^{i+1}}\times\frac{W}{2^{i+1}})$, where $i \in {1, 2, 3, 4}$. To integrate these features, the sequence is reshaped to $(C_i,\frac{H}{2^{i+1}},\frac{W}{2^{i+1}})$ and concatenated with the feature map. The concatenated features are fused using a pointwise convolution layer. In summary, CEF complements the TE Layer's output with the CE Layer's output, recovering local spatial information and enhancing fine-scale details. The output of all CEF layers is passed through a lightweight decoder from SegFormer to predict the final mask.





\begin{figure*}[t]
  \centering
  \noindent\includegraphics[width=0.96\linewidth]{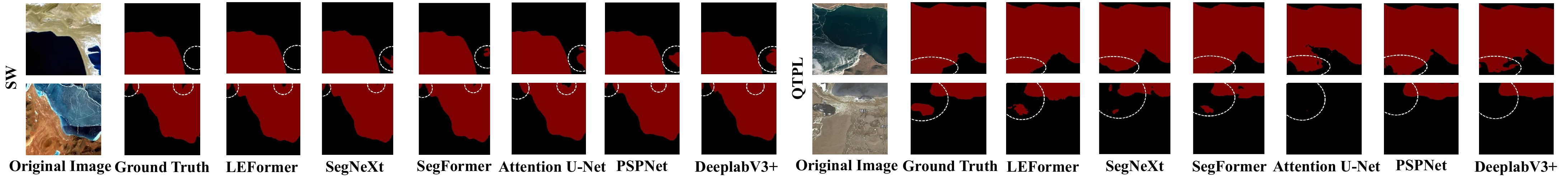}
  \caption{Visualization results of our proposed LEFormer and other methods. The white circles indicate apparent differences.}
  \label{fig:control_experiment}
\end{figure*}

\section{Experiments}

\subsection{Experimental Settings}

\textbf{Datasets}. This study evaluates the LEFormer's performance and generalization capability on two publicly satellite remote sensing datasets for lake extraction: the Surface Water dataset (SW dataset) and the Qinghai-Tibet Plateau Lake dataset (QTPL dataset)~\cite{QTPL-Dataset}. Both datasets include annotated lake water bodies from visible spectrum remote sensing images of size 256 × 256. The SW dataset comprises 17,596 images divided by 4:1 for training and testing, while the QTPL dataset contains 6,773 images divided by 9:1 for training and testing.
\\
\textbf{Implementation details}. The models are trained on a single Tesla V100 GPU using the \textit{MMSegmentation} codebase for 160K iterations on the SW and QTPL datasets. Data augmentation, such as random resizing and horizontal flipping, to enhance generalization. The AdamW~\cite{adamW} optimizer and cross-entropy loss function~\cite{cross-entropy} are used with a batch size of 16. The initial learning rate and weight decay are set to $6\times10^{-5}$ and 0.01, respectively. The learning rate is dynamically adjusted using a PolyScheduler~\cite{maskrcnn_2017} with a factor of 1.0. The models' accuracy is evaluated using overall accuracy (OA), F1, and mean Intersection over Union (mIoU). The efficiency is assessed by the parameter (Params, M) and floating point operations per second (Flops, G), denoted as \#P and \#F, respectively, in the table for brevity.

\begin{table}[t]
\centering
\caption{Impact of the number of PTLs on model parameters, flops, and accuracy (L denotes the number of layers).}
\fontsize{7.8}{9}\selectfont
\setlength{\tabcolsep}{4pt}
\begin{tabular}{c|c|c|ccc|ccc}
\hline
\multirow{2}{*}{L} & \multirow{2}{*}{\#P$\downarrow$} & \multirow{2}{*}{\#F$\downarrow$} & \multicolumn{3}{c|}{SW}                                                                             & \multicolumn{3}{c}{QTPL}                                                                            \\ \cline{4-9} 
                   &                            &                           & OA$\uparrow$                              & F1$\uparrow$                        & mIoU$\uparrow$                            & OA$\uparrow$                              & F1$\uparrow$                        & mIoU$\uparrow$                            \\ \hline
4                  & \textbf{3.24}              & \textbf{1.23}             & 94.68 & 92.35 & 88.98 & 98.39 & 98.03 & 96.72 \\
3                  & 3.26                       & 1.24                      & 95.04 & 92.88 & 89.69 & 98.64 & 98.33 & 97.22 \\
2                  & 3.48                       & 1.25                      & 95.52 & 93.57 & 90.64 & 98.66 & 98.36 & 97.26 \\
1                  & 3.61                       & 1.27                      & \textbf{95.63}                  & \textbf{93.73}                  & \textbf{90.86}                  & \textbf{98.74}                  & \textbf{98.45}                  & \textbf{97.42}                  \\ 
0                  & 3.74                       & 1.28                      & 95.62                  & 93.72                  & 90.84                  & 98.71                  & 98.42                  & 97.36                  \\ \hline
\end{tabular}
\label{table:ablation_study_a}
\end{table}

\begin{table}[t]
\centering
\caption{Impact of different encoder combinations on \#P, \#F, and accuracy (ETL + PTL denotes three ETLs and one PTL).}
\setlength{\tabcolsep}{3.5pt}
\fontsize{7.8}{9}\selectfont
\begin{tabular}{c|cc|cc|c|c|c|c|c}
\hline
\multirow{2}{*}{Dataset} & \multicolumn{2}{c|}{CE}     & \multicolumn{2}{c|}{TE}     & \multirow{2}{*}{\#P$\downarrow$} & \multirow{2}{*}{\#F$\downarrow$} & \multirow{2}{*}{OA$\uparrow$}              & \multirow{2}{*}{F1$\uparrow$}        & \multirow{2}{*}{mIoU$\uparrow$}            \\ \cline{2-5}
                         & DW           & MSCA         & ETL    & PTL      &                            &                           &                                  &                                  &                                  \\ \hline
\multirow{6}{*}{SW}      & \XSolidBrush & \XSolidBrush & \Checkmark   & \XSolidBrush & 3.22                       & 1.09                      & 95.60  & 93.68  & 90.79  \\
                         & \XSolidBrush & \XSolidBrush & \Checkmark   & \Checkmark   & 3.09                       & 1.07                      & 95.41  & 93.40  & 90.41 \\
                         & \Checkmark   & \Checkmark   & \XSolidBrush & \XSolidBrush & \textbf{0.74}              & \textbf{0.82}             & 94.32 & 91.84 & 88.28 \\
                         & \Checkmark   & \XSolidBrush & \Checkmark   & \Checkmark   & 3.46                       & 1.23                      & 95.55 & 93.60 & 90.69 \\
                         & \Checkmark   & \Checkmark   & \Checkmark   & \XSolidBrush & 3.74                       & 1.28                      & 95.62 & 93.72 & 90.84 \\
                         & \Checkmark   & \Checkmark   & \Checkmark   & \Checkmark   & 3.61                       & 1.27                      & \textbf{95.63}                   & \textbf{93.73}                   & \textbf{90.86}                   \\ \hline
\multirow{6}{*}{QTPL}    & \XSolidBrush & \XSolidBrush & \Checkmark   & \XSolidBrush & 3.22                       & 1.09                      & 98.65 & 98.35 & 97.25 \\
                         & \XSolidBrush & \XSolidBrush & \Checkmark   & \Checkmark   & 3.09                       & 1.07                      & 98.62 & 98.31 & 97.18 \\
                         & \Checkmark   & \Checkmark   & \XSolidBrush & \XSolidBrush & \textbf{0.74}              & \textbf{0.82}             & 98.36 & 97.99 & 96.66 \\
                         & \Checkmark   & \XSolidBrush & \Checkmark   & \Checkmark   & 3.46                       & 1.23                      & 98.64 & 98.34 & 97.22 \\
                         & \Checkmark   & \Checkmark   & \Checkmark   & \XSolidBrush & 3.74                       & 1.28                      & 98.71 & 98.42 & 97.36 \\
                         & \Checkmark   & \Checkmark   & \Checkmark   & \Checkmark   & 3.61                       & 1.27                      & \textbf{98.74}                   & \textbf{98.45}                   & \textbf{97.42}                   \\ \hline
\end{tabular}
\label{table:ablation_study_b}
\end{table}

\subsection{Ablation Studies}


Ablation studies are conducted to evaluate the impact of the number of PTLs on the model's performance on the SW and QTPL datasets. The results (Table \ref{table:ablation_study_a}) demonstrate that the highest accuracy is achieved with $L = 1$ (one PTL and three ETLs). This suggests that the PTL can be effectively utilized when the model suits the dataset. However, excessive use of the PTL can lead to reduced accuracy. 

We also conduct ablation studies to assess different encoder combinations' impact on performance and efficiency based on Table \ref{table:ablation_study_b}. TE with ETL and PTL achieves 90.41\% and 97.18\% mIoU on SW and QTPL. In contrast, CE with MSCA achieves 88.28\% and 96.66\% mIoU, indicating TE effectively captures long-range dependencies and global context. Additionally, including DW improves mIoU by 0.28\% and 0.04\%. MSCA improves mIoU by 0.17\% and 0.20\%, suggesting DW and MSCA contribute to accuracy. Omitting CE, PTL reduces mIoU by 0.38\% and 0.07\%. However, with DW and MSCA, PTL does not decrease metrics and results in a lighter model.

\begin{table}[t]
\centering
\fontsize{7.6}{9.1}\selectfont
\caption{Quantitative comparison of our LEFormer and other methods on the SW and QTPL datasets.}
\setlength{\tabcolsep}{2.1pt}
\begin{tabular}{c|c|c|ccc}
\hline
\multirow{2}{*}{Method}               & \multirow{2}{*}{\#P$\downarrow$} & \multirow{2}{*}{\#F$\downarrow$} & \multicolumn{3}{c}{SW / QTPL}                      \\ \cline{4-6} 
                                      &                                         &                                        & OA$\uparrow$ & F1$\uparrow$ & mIoU$\uparrow$ \\ \hline
PSPNet~\cite{PSPnet}                                                    & 46.80                      & 46.11                     & 94.17 / 98.40          & 91.73 / 98.03          & 88.12 / 96.77            \\
DeeplabV3+~\cite{DeepLabV3+}                                                & 54.70                      & 20.76                     & 94.30 / 98.39          & 91.87 / 98.03          & 88.28 / 96.75            \\
Attention U-Net~\cite{Attention-Unet} & 34.90                      & 66.64                     & 93.34 / 98.24          & 90.56 / 97.87          & 86.54 / 96.42                 \\
LANet~\cite{LANet}                                                     & 24.00                      & 8.31                      & 94.14 / 98.29          & 91.62 / 97.89          & 87.96 / 96.52             \\
SegFormer~\cite{SegFormer} & 3.72                      & 1.59                     & 95.58 / 98.66          & 93.65 / 98.36          & 90.75 / 97.27                 \\
SegNeXt~\cite{SegNeXt}                                                     & 4.26                      & 1.55                      & 95.50 / 98.60          & 93.56 / 98.30          & 90.61 / 97.15\\
MSNANet~\cite{MSNANet}                                                   & 72.30                      & 61.94                     & 94.44 / 98.47          & 92.12 / 98.11         & 88.66 / 96.88              \\
\textbf{LEFormer (Ours)}                                   & \textbf{3.61}              & \textbf{1.27}             & \textbf{95.63 / 98.74} & \textbf{93.73 / 98.45} & \textbf{90.86 / 97.42}  \\ \hline
\end{tabular}
\label{table:results_sw}
\end{table}


\subsection{Comparison to the State-of-the-Arts}

We assess the efficacy of LEFormer by comparing it with advanced lake extraction models~\cite{MSNANet, SegNeXt, SegFormer, LANet, Attention-Unet, DeepLabV3+, PSPnet} on the SW and QTPL datasets. Table \ref{table:results_sw} summarizes the quantitative results and visualization results in Fig. \ref{fig:control_experiment}.

As shown in Table \ref{table:results_sw}, LEFormer outperforms all other models regarding parameters, flops, and accuracy. On the SW dataset, LEFormer achieves 90.86\% mIoU using only 3.61M parameters and 1.27G flops. Compared to the previous lake extraction method MSNANet, LEFormer is 20$\times$ minor and requires 48$\times$ fewer flops while achieving a 2.20\% better mIoU. On the QTPL dataset, LEFormer also achieves SOTA performance. In summary, these results demonstrate the superiority of LEFormer in the lake extraction task.

\section{Conclusion}
In this study, we propose the LEFormer, a hybrid CNN-Transformer architecture for accurate lake extraction. We combine CNNs and Transformers to capture short- and long-range dependencies to obtain robust features for lake mask prediction. Experiments show that LEFormer achieves SOTA performance and efficiency on two benchmark datasets. We hope that LEFormer will encourage the design of efficient hybrid CNN-Transformer networks for lake extraction.

\section{Acknowledgements}
This work is supported in part by the Natural Science Foundation of China under Grant No. 62222606 and 62076238.



\bibliographystyle{IEEEbib}
\bibliography{strings,refs}

\end{document}